\documentclass[letterpaper, 10 pt, conference]{ieeeconf}  %

\IEEEoverridecommandlockouts                              %

\usepackage{dsfont}
\usepackage{amsmath} %
\usepackage{amssymb}  %
\usepackage{hyperref}
\hypersetup{
    colorlinks=true,
    linkcolor=blue,
    filecolor=magenta,      
    urlcolor=cyan,
}
\usepackage{graphicx}
\graphicspath{ {img/} }
\usepackage{caption}
\usepackage{subcaption}
\usepackage{ dsfont }

\usepackage[T1]{fontenc}
\usepackage[utf8]{inputenc}
\usepackage[ruled,vlined]{algorithm2e}
\usepackage{algorithmic}
\newcommand{\prim}{{\fontfamily{qcr}\selectfont
PickAndPlace }}
\usepackage{wrapfig}
\usepackage{makecell}

\usepackage{xcolor}

\DeclareMathOperator*{\argmin}{arg\,min}
\DeclareMathOperator*{\argmax}{arg\,max}

\usepackage{booktabs}

\newcounter{mycounter}  
\newenvironment{noindlist}
 {\begin{list}{\arabic{mycounter}.~~}{\usecounter{mycounter} \labelsep=0em \labelwidth=0em \leftmargin=0em \itemindent=0em}}
 {\end{list}}

\title{\LARGE \bf
Efficient and Interpretable Robot Manipulation with \\ Graph Neural Networks
}

\author{Yixin Lin$^{*1}$, Austin S. Wang$^{1}$, Eric Undersander$^{1}$,  and Akshara Rai$^{*1}$%
\thanks{$^{*}$ Equal contribution}%
\thanks{$^{1}$ The authors are with Facebook AI Research, in Menlo Park, CA.
        {\tt \footnotesize yixinlin,wangaustin,eundersander,akshararai@fb.com}}%
}

\begin{document}

\maketitle
\thispagestyle{empty}
\pagestyle{empty}

\begin{abstract}

Manipulation tasks, like loading a dishwasher, can be seen as a sequence of spatial constraints and relationships between different objects.
We aim to discover these rules from demonstrations by posing manipulation as a classification problem over a graph, whose nodes represent task-relevant entities like objects and goals, and present a graph neural network (GNN) policy architecture for solving this problem from demonstrations.
In our experiments, a single GNN policy trained using imitation learning (IL) on 20 expert demos can solve blockstacking, rearrangement, and dishwasher loading tasks; once the policy has learned the spatial structure, it can generalize to a larger number of objects, goal configurations, and from simulation to the real world.
These experiments show that graphical IL can solve complex long-horizon manipulation problems without requiring detailed task descriptions.
Videos can be found at: \url{https://youtu.be/POxaTDAj7aY}.
\end{abstract}

\section{Introduction}

\begin{figure*}
    \centering
    \vspace{0.1cm}
    \includegraphics[width=0.9\textwidth]{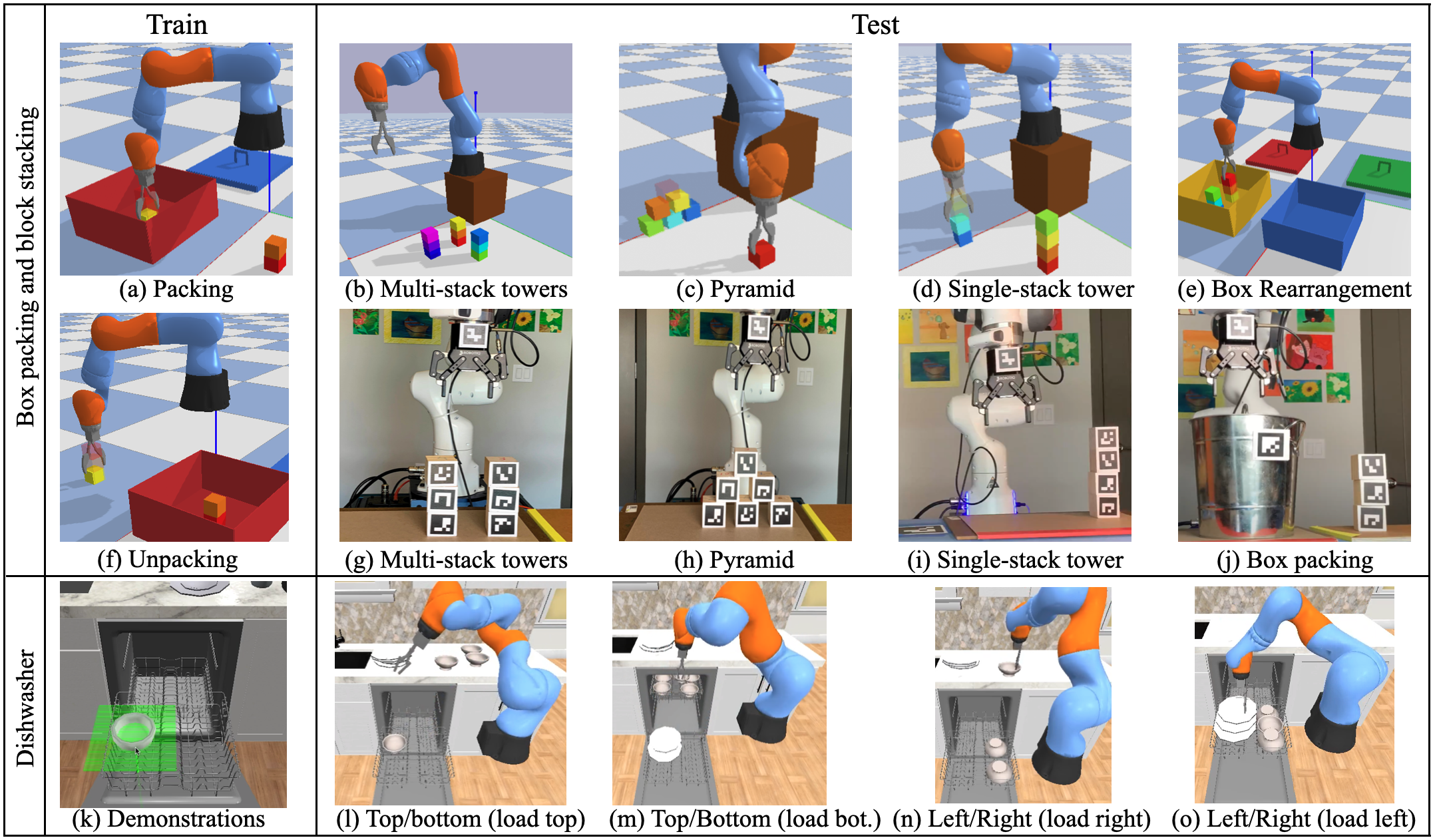}
    \caption{\small We train a policy on small instances of the problem (left column: (a), (f), (k)) and test generalization on new, larger problem instances in both simulation ((b)-(e)), and on real hardware ((g)-(j)). We also apply the method to a complex dishwasher-loading environment, generating training data using a point-and-click interface ((k)) and testing on a variety of scenarios ((l)-(o)), described more fully in \ref{sec:dishwasher}.}
    \label{fig:intro}
    \vspace{-0.5cm}
\end{figure*}

Everyday manipulation tasks deal with relationships and constraints between objects and environments. For example, loading a bowl in a dishwasher requires pre-conditions, like an open dishwasher and a grasped bowl. %
Specifying such pre-conditions  for complex tasks can be tedious and error-prone. In the above example, the bowl won't get cleaned if it is placed in the dishwasher in the wrong orientation. %
Consider a scenario with a user and her personal robot. %
The user wants to teach the robot her preferred method of loading a dishwasher -- bowls on the top and plates at the bottom. Typical Task and Motion Planning (TAMP) would require the user to write a detailed symbolic task and goal description, which can can be cumbersome for non-experts. It is easier for the user to demonstrate her preference by simply loading the dishwasher with a few plates and bowls. Moreover, once the user has demonstrated her preference with a few plates and bowls, the robot should generalize the instructions to \textit{any} number of plates and bowls. This is the central problem that we address -- how do we learn task structure from very few demonstrations and then generalize this knowledge to arbitrary numbers of objects, as well as other related tasks? 

We hypothesize that user demonstrations of successful task completion inherently contain task-specific rules. Given an appropriate state representation, imitating user's actions in a particular state is enough for successful task completion. In the dishwasher scenario, the user would flip the bowl before loading, encoding her preference. We aim to learn such task-specific rules by representing the environment state as a graph whose nodes represent task-relevant entities like objects (plates, bowls, dishwasher) and target positions (or goals) of objects. Next, we train a graph neural network (GNN) that operates over this graph and selects the most relevant object in the scene, a suitable goal state for the selected object, and an action that achieves this transition. The GNN policy architecture enables generalization over variable number of objects in the scene, as GNNs are invariant to the number of nodes in a graph. %
Our experiments show that a trained GNN policy generalizes to tasks of increased complexity and variable numbers of objects, starting with as few as 20 expert demonstrations (Fig. \ref{fig:intro}). Additionally, we extract interpretable explanations from GNNs, by modifying \cite{ying2019gnnexplainer}. We find the nodes and features that were most important for the decision made by the GNN policy at a current state, giving interpretable explanations like `object $i$ was chosen because of its neighbor $j$ and feature $z$'. Using this explainer, we can identify if a GNN is overfitting, as well as verify that the GNN indeed learns the task structure (Section \ref{sec:gated_explainer}).

Our approach depends on a hierarchical decomposition of manipulation tasks that can reproduce the expert demonstrations well. %
We assume known robot-specific primitives like \prim and learn a GNN policy that provides inputs like desired positions and orientations to these primitives. 
This hierarchical setup has several advantages: (1) sample-efficient learning; our GNN policy can train from 20 expert demonstrations. (2) Minimizing supervised learning loss on expert demonstrations can solve complex tasks, without explicitly specifying the spatial constraints of the task. (3) Transfer of learned high-level task policies \textit{across morphologies}. For example, in Sec. \ref{sec:dishwasher} we present experiments where a dishwasher loading policy is learned from 5 human point-and-click demonstrations and applied to a simulated robot loading a dishwasher.%

The main contributions of our work are presenting
(1) GNNs as a promising policy architecture for long-term manipulation tasks,
(2) imitation learning as a well-suited training scheme for such a policy choice, and
(3) a modified GNNExplainer to interpret the decisions made by our learned policy.
We conduct experiments on a Franka arm in the real world and in two simulated environments - a dishwasher environment and a blockstacking and box rearrangement environment (Figure \ref{fig:intro}). In the dishwasher environment, the robot loads a dishwasher with plates and bowls; in the box rearrangement setting the robot moves blocks from one box to another. On hardware, the robot stacks blocks in different goal configurations, and places blocks in a bucket. We train GNN policies that can achieve these tasks starting from a small set of expert demonstrations (5 in dishwasher loading and 20 in box rearrangement). We compare our approach against reinforcement learning (RL) with both feedforward NN and GNN and show that imitation learning on GNN outperforms traditional learning-based approaches. %

\section{Related Work}
\subsection{Graphical approaches to manipulation}
Graphical representations of scenes have been used for learning high-dimensional dynamics models \cite{fragkiadaki2016learning, ye2019objectcentric}, learning object-relevance in problems with large object instances \cite{silver2020planning}, visual imitation learning \cite{sieb2020graph, huang2019neural}, and high-level policies \cite{li2020towards}.  
\cite{huang2019neural} propose a Neural Task Graph (NTG) that use a graph to represent the action sequence of a task. \cite{silver2020planning} train a GNN to predict if a particular object in a scene is relevant to the planning problem at hand. %
\cite{sieb2020graph} tackle \textit{visual} imitation learning through video demonstrations, while we assume access to more state information but increase task complexity and demonstrate strong zero-shot generalization on increased object counts. Unlike \cite{li2020towards} which focuses on training GNNs using RL, we use imitation learning on comparatively tiny amount of expert demonstration data and show generalization to tasks beyond \cite{li2020towards} (Section \ref{sec:exp}).
\cite{kapelyukh2021my} also uses GNNs and focus on predicting preference-conditioned object goal poses in continuous space. However, they do not attempt to tackle the sequential decision-making problem, instead predicting all the desired poses at once.

\subsection{Task and motion planning (TAMP)}

TAMP is a powerful tool for solving long-horizon manipulation tasks, combining discrete symbolic task planning with continuous motion planning. We refer readers to \cite{garrett2020integrated, kaelbling2011hierarchical} for an overview. TAMP algorithms rely on pre-defined symbolic rules, or \textit{planning domains}, defining the state, actions, transition models (effects), and constraints that are used by symbolic planners \cite{konidaris2018skills,fox2003pddl2, holler2020hddl, gharbi2015combining}.  Given a domain definition for a task, TAMP can deal with arbitrary numbers of objects, in any configuration, occlusions and partial observability. 
However, domains can be hard to define in complex environments like dishwasher loading. 
Apart from different actions, like picking plates and bowls in different orientations, pulling and pushing trays, and their effects, there are many feasibility conditions that need to be specified in the domain. For example, a tray can only be loaded if it is open; bottom tray can only be loaded if it is open, but the top tray is closed; any tray can only be loaded if there are empty slots. Similarly user preferences about the orientation and position of dishes in the dishwasher need to be symbolically specified. Once the domain is defined for one goal configuration, switching to a different desired goal requires editing the domain. While experts are able to design and maintain TAMP domains, non-expert users can find this challenging. We aim to simplify domain design in TAMP without losing its generalizability. 

Incorporating learning in TAMP is a popular area of research, but most works assume a known symbolic task description, used with a planner. Given the task plan, they learn low-level skills \cite{szot2021habitat}, or parameters of the low-level skills \cite{wang2021learning}, or transition models for the skills \cite{loula2020learning}. Learning has been successful in speeding up planning in TAMP \cite{kim2018learning, wells2019learning, chitnis2016guided, driess2020deep, kim2020learning}; \cite{wang2018active, kaelbling2017learning} learn transition models over symbolic states and actions, eliminating the need for hand-crafted transition tables. In contrast, we do not decompose our problem into learning models or constraints, followed by planning. Instead, we directly learn a policy using imitation learning, and achieve generalization through the choice of our graphical state and policy representation. This circumvents the need to define a symbolic description of the task, while maintaining other advantages of TAMP, like generalization to any number of objects. Specifically, we use imitation learning to train a high-level policy that operates on pre-defined low-level skills to achieve new, unseen tasks. Our policy implicitly learns about the feasibility domain (e.g. only picking the top block in a stack) while generalizing to solve unseen tasks (e.g. stacking multiple towers).

\section{Background}
\subsection{Reinforcement learning and imitation learning}

We consider a Markov Decision Process (MDP) with a continuous state space $\mathcal{S}$ and a high-level discrete action space $\mathcal{A}$.
Starting from state $s_t$, executing high-level action $a_t$ incurs a reward $r_t$ and leads to state $s_{t+1} \sim p( s_{t+1} | s_t, a_t)$ following the transition function $p$.
Given this problem setup, we aim to learn a policy $\pi_\theta(s_t) = a_t$ that imitates an expert demonstration.
For an expert dataset of $N$ trajectories $\mathcal{D} = \{ \tau_i \}_{i=1}^N, \tau_i = \{ s_{i, 1}, a^{exp}_{i, 1}, s_{i, 2}, a^{exp}_{i, 2}, \dots, s_{i, T}, a^{exp}_{i, T} \}$, we minimize the supervised learning loss: $\min_\theta \mathbb{E} [ \sum_{i=1}^N \sum_{t=1}^T \lVert a^{exp}_{i,t} - a^{pred}_t \rVert]$,
where $ a^{pred}_t = \pi_\theta(s_{i, t})$. %
Our graphical state and policy representations and induced inductive biases generalize outside of the training distribution of expert demonstrations.

\begin{figure}[t]
    \centering
    \includegraphics[width=0.5\textwidth]{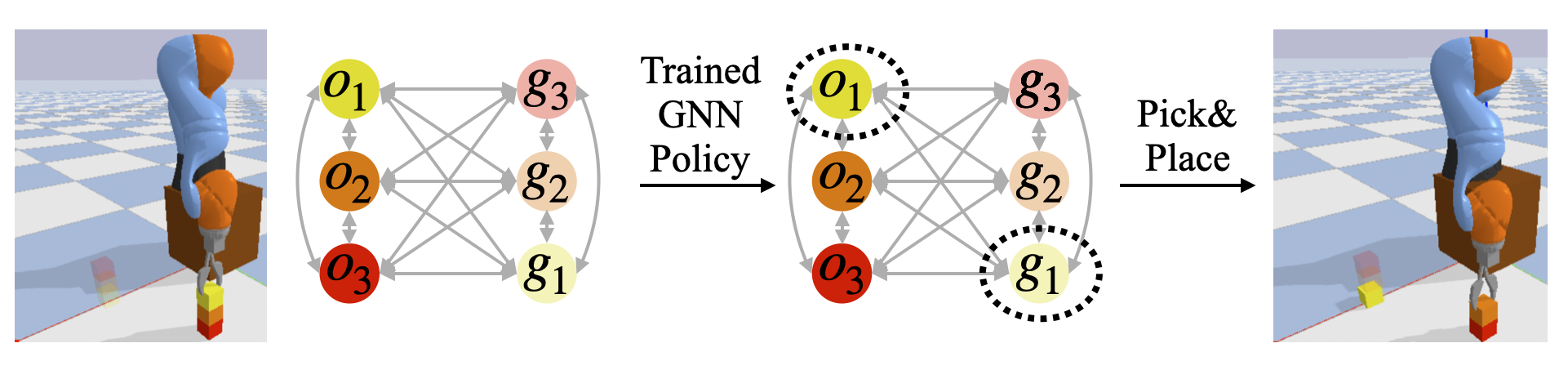}
    \caption{\small An overview of our approach. We train a high-level GNN policy that takes a graph representation of state as input and selects the next object to pick, and the next goal to place it in. A low-level \prim primitive then picks the chosen object and places it in the desired goal. Summary in Algo \ref{algo:GNN_algo}.}
    \label{fig:overview}
    \vspace{-0.5cm}
\end{figure}
\begin{algorithm}[t]
\caption{\footnotesize Long-horizon manipulation with GNNs}
\begin{algorithmic}
    \footnotesize

    \STATE Given a graph dataset $D$ of $N$ expert demonstrations\\
    
    \STATE Randomly initialize a GNN policy $\pi_\theta$\\
    \For{each gradient step}{
    \STATE Update $\theta^* = \argmin_{\theta} \mathcal{L}(\theta, D)$ \\
    \STATE where $\mathcal{L}$ is the cross-entropy loss in Eq. \ref{eq:ce}
    }
    \For{each time step}{
        \STATE Create graph $G$ from the environment state \\
        \STATE Choose $o, g = \pi_\theta (G)$ \\
        \STATE Execute \prim ($o \rightarrow g$)
    }
\end{algorithmic}
\label{algo:GNN_algo}
\end{algorithm}
\setlength{\textfloatsep}{0pt}%

\subsection{Graph Neural Networks (GNNs)}
\label{sec:gnn-architectures}

GNNs \cite{battaglia2018relational} are deep networks designed to operate on graphs.
Let $G$ be a graph with nodes $V$ and undirected edges $E$, where each node $v \in V$ is associated with a feature vector $\phi(v)$.
A single \textit{message-passing} GNN layer applies a message-passing function on every node, updating each node's feature as a function of its own and its neighbors' features; a GNN model commonly stacks multiple layers.
At each layer $l$ and for every node $v_i \in V$, we update the node's feature vector $h_i^l = f_\theta^l(h_i^{l-1}, \{h_j^{l-1}\}_{j \in \mathcal{N}_i})$, where $h_i^l$ is the updated node feature and $h_i^0 = \phi(v_i)$ is the input feature.
$f_\theta$ is a parametrized function whose weights $\theta$ are learned using gradient descent during training.
 $f$ and $\theta$ are \textit{shared} across all nodes; once the parameters $\theta$ are learned, the GNN can be applied to a new graph with any number of nodes.
GNNs are highly parallelizable and efficient to compute; we use Pytorch Geometric \cite{fey2019fast,paszke2019pytorch} for all our computations.

Different GNN architectures make different choices of $f_\theta$ that induce different inductive biases on the problem at hand. We experiment with four kinds of GNN architectures:

\noindent \textbf{Graph Convolution Networks (GCN): } GCNs \cite{morris2019weisfeiler} are isotropic graph networks where each neighbour's contribution is weighed by the edge weight of the connecting edge: 
$h_i^l = \sigma(\theta_1 h_i^{l-1} + \theta_2 \sum_{j \in \mathcal{N}(i)} e_{j,i} \cdot h_j^{l-1} $).

$\theta_1$ and $\theta_2$ constitute the learnable parameters, $\sigma$ is the activation function, such as the ReLU activation.

\noindent \textbf{GraphSage (Sage): } GraphSage \cite{hamilton2017inductive} is also an isotropic network like GCNs that takes the mean features of each of its neighbors without taking edge weights into account: 

$ h_i^{l} = \sigma(\theta_1 h_i^{l-1} + \frac{\theta_2}{|\mathcal{N}(i)|}  \sum_{j \in \mathcal{N}(i)} h_j^{l-1}) $. 

\noindent \textbf{GatedGCN (Gated): } GatedGCN \cite{li2015gated} is an anisotropic graph convolution network, where the weights on the neighbors are learned using a Gated Recurrent Unit (GRU).  

$ h_i^{l} = \text{GRU}( h_i^{l-1} , \sum_{j \in \mathcal{N}(i)} \theta_1 h_j^{l-1})$

\noindent \textbf{Graph Attention Networks (Attention): } Graph attention networks \cite{velivckovic2017graph} are anisotropic graph convolution networks that learn relative weights between neighbors using an attention mechanism: $h_i^l = \sigma( \theta_1 h_i^{l-1} + \sum_{j \in \mathcal{N}(i)} a_{i,j} \theta_2 h_j^{l-1}) $,
    where a learned self-attention weight $a_{i,j}$ measures the strength of connection between nodes $v_i$ and $v_j$.

\section{GNN policies for manipulation}

In this section, we explain our formulation which casts manipulation tasks as operations over a graph.
We assume a low-level \prim primitive which, given an object and a goal, grasps the chosen object and places it in the desired goal. 
We train a high-level GNN policy that takes a graph representation of environment as input and selects the block and goal location input to {\fontfamily{qcr}\selectfont PickAndPlace}. The dishwasher loading policy additionally predicts target pick and place orientations, and also chooses which action to use, like {\fontfamily{qcr}\selectfont OpenTray} for opening a dishwasher tray. For clarity, we will describe the next section using only \prim and leave additional details about dishwasher loading to Section \ref{sec:dishwasher}.    
Our approach is outlined in Fig. \ref{fig:overview}, Algo. \ref{algo:GNN_algo}.

\begin{figure*}
    \vspace{0.25cm}
    \centering
    \includegraphics[width=\textwidth]{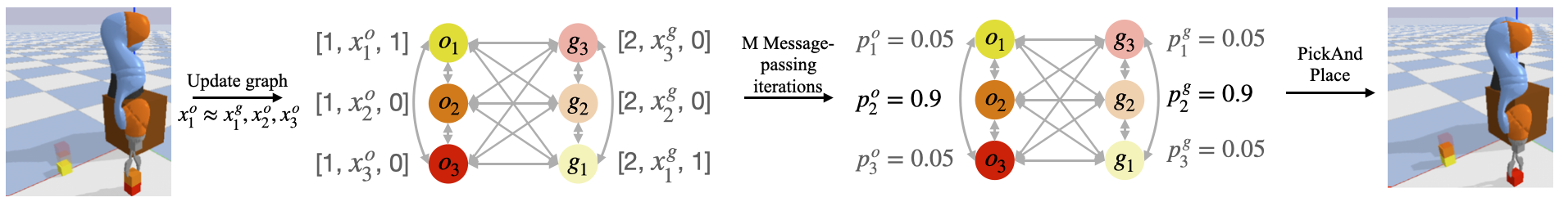}
    \caption{\small Overview of our algorithm at a timestep. Our method takes in an observation, transforms it into a graph with a 5-dimensional feature per node, and passes it to the GNN policy, which selects an object and goal to input to \prim.}
    \label{fig:visualization}
    \vspace{-0.6cm}
\end{figure*}

\subsection{Problem formulation: Graphical representation of state}
\label{sec:gnn_setup}

We encode the scene as a graph, whose nodes consist of the task-relevant entities, such as objects and their target positions (goals).
Let there be $K$ objects, and $L$ goals in the scene.
We create a graph $G = (V, E)$, where the vertices $V = \{ v^o_k \}_{k=1}^K \cup \{ v^g_l \}_{l=1}^L$ represent the objects and goals in the scene,
giving us a total of $K+L$ nodes.
We create a dense, fully-connected graph, where all nodes are connected to all other nodes; $E = \{ e_{i,j} \}$ for $i,j = 1 \dots K+L$.

Each node $v \in V$ in the graph has a feature vector $\phi(v)$, which contains node-specific information.
The input features of each node are 5-dimensional: a categorical feature $\{0,1,2,3\}$ denoting if a node is a cover, block, goal for a block, or goal for a cover, the 3-dimensional position of the object or goal in the frame of the robot, and a binary feature which is 1 if a goal is filled or an object is in a goal, and 0 for empty goals or objects.
The current state graph is input to the GNN policy, which outputs a categorical distribution over objects and goals. 
The selected object and goal positions are sent as inputs to the \prim primitive.
This is illustrated in Figure \ref{fig:visualization} for a $K=L=3$ block stacking trajectory. Our approach directly generalizes to situations where number of goals and objects are different. For example, in box rearrangment, the policy learns to move the box cover \textit{out of the way} by placing the cover on the table before moving blocks, and finally closing the box.

In this work, we deal with problems with a shared underlying task structure -- for example, pick the highest block from a stack, and place it in the lowest free goal. We use expert demonstrations to train a GNN policy which learns this underlying structure, in contrast to traditional TAMP, where such constraints are pre-defined. Once this structure is learned, the policy automatically generalizes to new unseen problems, as long as the underlying task structure holds. %

\subsection{Training the GNN from demonstrations}

We pose a long-horizon manipulation problem as a classification problem at each high-level step where a decision is made over \textit{which} object to move to \textit{where} using \textit{what} action.
The output of the GNN policy is $K+L$ dimensional, reshaped into $K$ and $L$ dimensional outputs
$V^\text{out}_g = \{ v^g_l \}_{l=1}^L$ and $V^\text{out}_o = \{ v^o_k \}_{k=1}^K$.
$V^\text{out}_o$ and is then passed through a softmax function $p(o_j) = \exp(v^o_j)/\sum_{k=1}^K \exp(v^o_k)$ to generate a $K$-dimensional categorical distribution $P^o_\text{pred} = \{ p^o_1, p^o_2, \cdots p^o_K \}$ depicting the picking probabilities of objects.
The object with the highest predicted probability $o_* = \argmax_{j} p(o_j)$ is the output of the GNN.
The same transformation is applied to goals, resulting in a probability distribution $P^g_\text{pred} = \{p^g_1, p^g_2, \cdots p^g_L \}$ and picked goal $g_*$. %

Given target distributions $P^o_\text{tgt}$ for the objects and $P^g_\text{tgt}$ for goals from expert data, the GNN policy parameters $\theta$ are trained to minimize the cross-entropy loss: 
\begin{align}
\label{eq:ce}
    \argmin_\theta \Big[ - \sum_{k=1}^K [P^o_\text{tgt}]_k \log (p^o_k) - \sum_{l=1}^L [P^g_\text{tgt}]_l \log(p^g_l) \Big]
\end{align}

The expert demonstrations used for training the GNN policy are also cast as a graph with target output distributions coming from the expert action. 
We collect $N$ demonstrations of the expert solving the task. 
At each step $t$, we extract input-output pairs $\{(s_t = (o_{k=1, \cdots, K}, g_{l=1, \cdots, L}), a_t) \}$, where $o_i$ and $g_i$ are the objects and goals in the scene, and $a_t = \{o^\text{exp}_{t}, g^\text{exp}_{t}\}$ is the action taken by the expert, indicating the next object $o^\text{exp}$ to be moved to the next goal $g^\text{exp}$.
This generates a training dataset $D$ of $N$ expert demonstrations solving multiple tasks: $D = \{\tau_n \}_{n=1}^N, \tau_n = \{ s_1, a_1, s_2, a_2, \dots, s_T\}$

For training the GNN policy, we sample a batch of state-action pairs from the dataset $D$ and convert each sampled state $s_b$ into a graph $G_b=(V_b,E_b)$, as described in Section \ref{sec:gnn_setup}. $s_b$ is also used to create the graph nodes $V_b$ and corresponding node features $\phi_b$, and edges $E_b$. 
The expert action $a_b = \{o^\text{exp}_{b}, g^\text{exp}_{b}\}$ is converted into two $K$ and $L$-dimensional 1-hot target distributions $P^o_\text{tgt}$ and $P^g_\text{tgt}$ for goal and object prediction. $P^o_\text{tgt} = \mathds{1}[ o_k  = o^\text{exp}_b]$ is a one-hot vector: 1 for the object chosen by the expert, and 0 for all others. %
Parameters $\theta$ of the GNN are learned to minimize the cross-entropy loss (Eq. \ref{eq:ce}) between prediction of the GNN policy given $G_b$ as input, and target distributions $P^o_\text{tgt}$ and $P^g_\text{tgt}$.

This high-level policy could be learned in many ways. For example, we could learn a multilayer perceptron (MLP) to predicts the next block and goal. 
However, if the MLP policy is trained on $K=3$ objects, it does not generalize to $K=4$, since the number of inputs, and architecture of the policy are different for different $K$.  On the other hand, GNNs generalize to different number of nodes in the graph, and hence can be used on variable number of objects.
Our GNN policy trained on $K=3,4$ shows zero-shot generalization up to $K=9$ (Section \ref{sec:rl_comparison}). 

\subsection{Interpreting the learned GNN policy}

\label{sec:explainer}

\cite{ying2019gnnexplainer} propose a GNNExplainer that adds interpretability to GNNs by determining importance of neighbouring nodes and input features for decision making. Intuitively, \cite{ying2019gnnexplainer} find a subgraph and subset of input features that result in the smallest shift in the output distribution of the GNN. We modify this GNNExplainer to suit our problem setting.%
We aim to find a mutated graph $G_S$ and feature mask $F$, such that the output of $\pi_\theta$ given $G_S$ and masked features $\phi_S = \phi \odot F$ is close to $P^o_\text{pred}, P^g_\text{pred}$. This setup is different from \cite{ying2019gnnexplainer} where a categorical distribution is predicted for \textit{every} node in a graph; our model instead predicts over \textit{all} nodes. As a result, the number of nodes in our mutated graph $G_S$ are the same as in $G$; this identifies which spatial relationship, or neighbours contributed most to the policy's decision. %

Given a trained GNN $\pi_\theta$ and input graph $G = (V,E)$, we aim to find a mutated graph $G_S = (V, E_S), E_S \subset E$ and a feature mask $F$, such that the mutual information between $Y = \pi_\theta(G, \phi)$, and $Y_S = \pi_\theta(G_S, \phi_S=\phi \odot F)$ is maximized:  
\vspace{-0.2cm}
\begin{align}
\label{eq:mi}
    G_S, F &= \argmax_{G_S, F} \text{MI}(Y, Y_S)
    = H(Y) - H(Y | Y_S) 
\end{align}
\vspace{-0.2cm}

$H(Y)$ does not depend on $G_S$ or $F$, %
hence maximizing the mutual information between $Y$ and $Y_S$ is the same as minimizing the conditional entropy $H(Y | \pi_\theta(G_S, \phi_S))$. Intuitively, the explanation for $Y$ is a mutated graph $G_S$ and feature mask $F$ %
that minimize the uncertainty over $Y$ 
\vspace{-0.2cm}
\begin{align}
\label{eq:red_mi}
    G_S, F = \argmin_{G_S, F}  H(Y | \pi_\theta(G_S, \phi_S))
\end{align}

\begin{figure}[t]
    \vspace{0.25cm}
    \centering
    \includegraphics[width=0.72\linewidth]{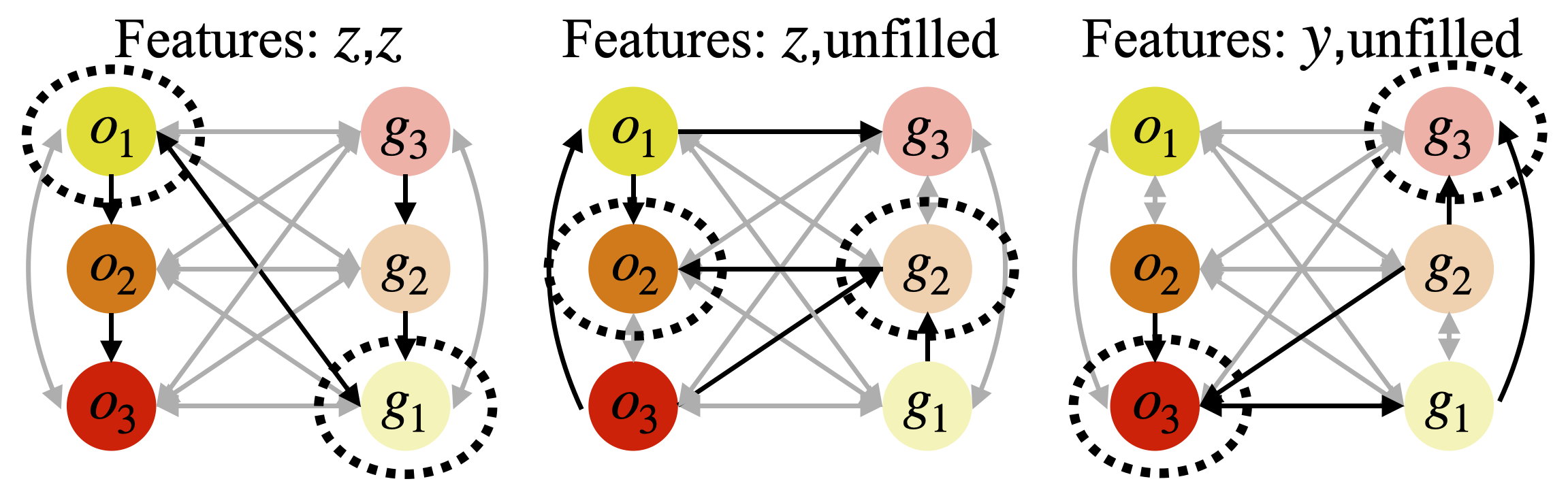}
    \caption{\small Visualizing the most important edges and features for choosing each block over a 3-step trajectory. The circled object and goal are the ones selected by the policy. The most important edge is bolded; the most important feature is listed by time step.}
    \label{fig:explainer_viz}
\end{figure}
We limit the total number of alive edges $|E_S| \leq c_E$, and alive features $\sum_j F_j \leq c_F$, where $c_E$ and $c_F$ are hyperparameters. %
Figure \ref{fig:explainer_viz} explains the GNN decisions on the 3-block environment, visualizing the 3 most important edges and feature.%
We extract interpretable explanations of the form ``node $i$ was chosen because of its relationship with nodes $j,k,l$; the most important feature was block height $z$".
The identified important edges always start or end on the \textit{selected object}, implying that the policy's decision was informed by how the selected block relates to its neighbours. %

\section{Experiments}
\label{sec:exp}

\begin{figure}[b]
\centering

    \begin{subfigure}{0.96\linewidth}
        \centering
        \includegraphics[width=0.24\linewidth,height=0.24\linewidth]{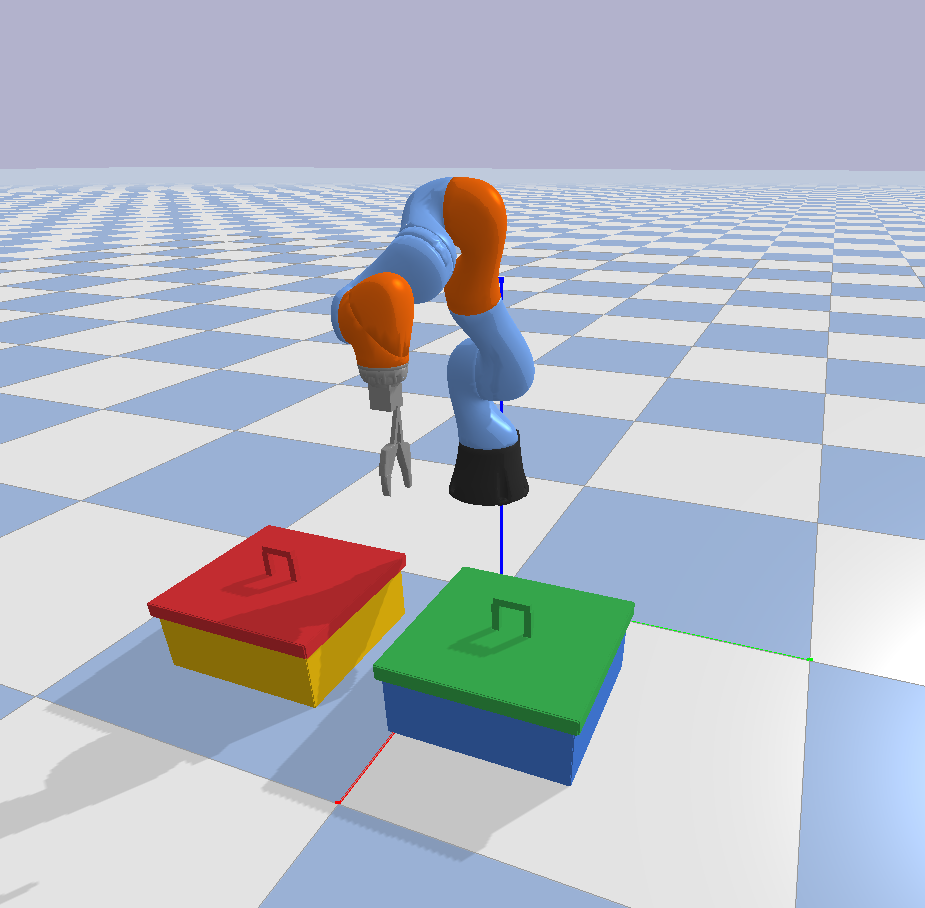}
        \includegraphics[width=0.24\linewidth,height=0.24\linewidth]{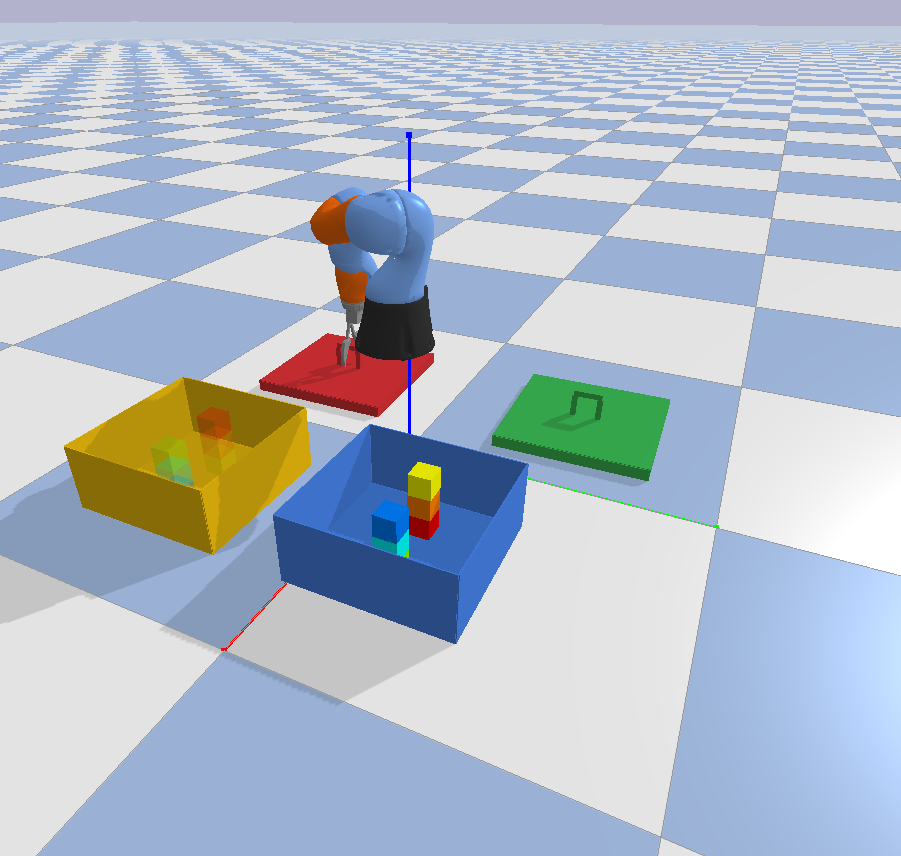}
        \includegraphics[width=0.24\linewidth,height=0.24\linewidth]{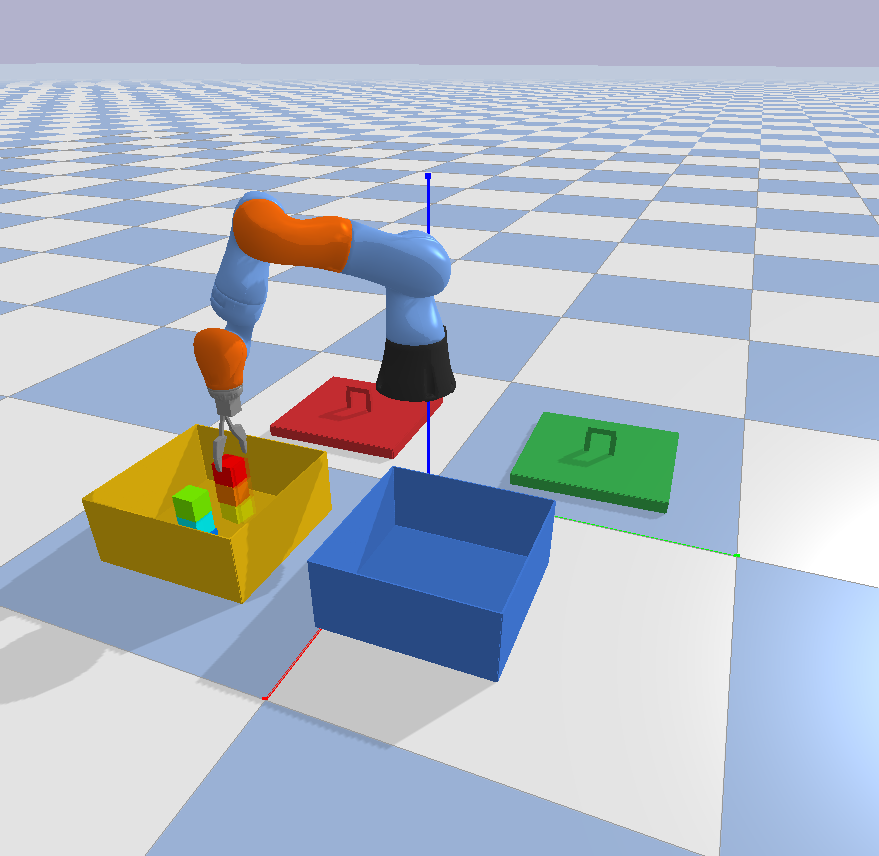}
        \includegraphics[width=0.24\linewidth,height=0.24\linewidth]{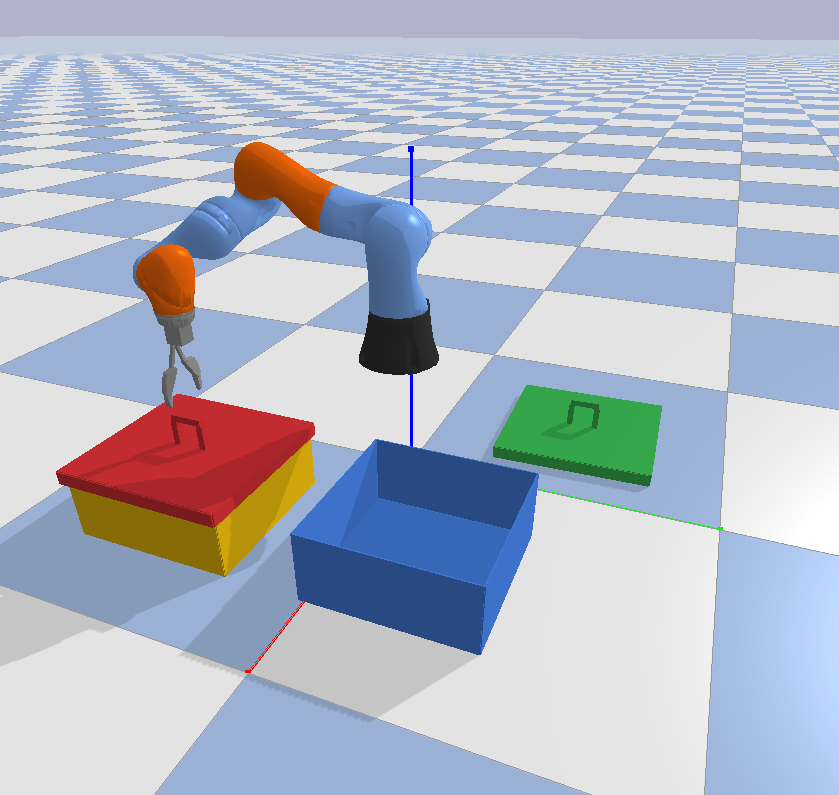}
    \end{subfigure}
    \caption{\small Stages of the rearrangement experiment.}
    \label{fig:rearrangement-steps}
\end{figure}

We use a Franka Panda manipulator equipped with a Robotiq 2F-85 two-finger gripper, and solve blockstacking and box packing tasks on hardware. For detecting blocks on hardware, we utilize a RealSense depth camera with the ArUco ARTags library \cite{garrido2016generation}. In simulation, we create two environments - dishwasher loading in AI Habitat \cite{szot2021habitat}, and blockstacking and box rearrangement in PyBullet\cite{coumans2016pybullet}. All train and test environments are shown in Figure \ref{fig:intro}.

\subsection{Block stacking and box packing experiments}

\label{sec:rl_comparison}

Each environment contains $K$ blocks, with different initial and goal positions. Success is measured by percentage of goals filled at the end of each trial. This experiment studies the generalization of the trained GNN policy across large number of blocks, multiple boxes and unseen tasks like pyramids and multiple stacks. Environments are in Fig \ref{fig:intro}a-j:

\begin{noindlist}

    \item \textbf{$K$-block stacking}: $K$ blocks are initialized in a random location; the goal is to invert them at another random location, demonstrating generalization to number of objects. %
    
    \item \textbf{$K$-pyramid}: same as $K$-block, but goal positions are in a pyramid configuration (Figure \ref{fig:intro}c), analyzing robustness to new goal configurations for the blocks.
    
    \item \textbf{$K$-block $s$-stack}: $s$ stacks of $K$ blocks  (Figure \ref{fig:intro}b), generalizing to variations in both initial and goal configurations.
    
    \item \textbf{Box Rearrangement}: This environment has two closed boxes, one empty, and the other containing $K$ blocks. The task is to open the boxes, move the blocks from one box to the other, and close the boxes. This task is different from our other blockstacking tasks, but uses the \textit{same} trained GNN policy. It tests robustness to partial observability and occlusion; the policy does not know location of blocks until the boxes are opened and has to move the covers to a ``storage" location before moving the blocks (Figure \ref{fig:rearrangement-steps}).  
   
\end{noindlist}

For all experiments, we consider 4 variants of our approach (\textbf{IL-GNN}), consisting of different GNN architectures (Sec. \ref{sec:gnn-architectures}).
All policies consist of 3 hidden layers, with 64 hidden units each and ReLU activation.
For attention policies, the number of attention heads were set to 1. 

\vspace{-0.2cm}

\begin{figure}[b]
    \centering
    \includegraphics[width=0.9\linewidth]{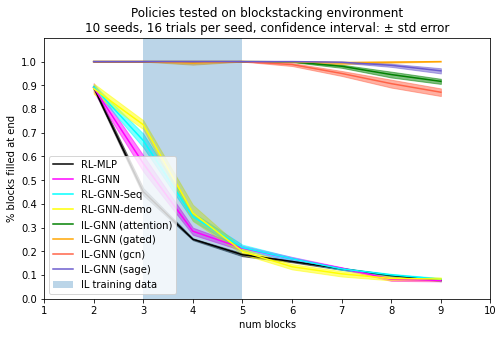}
    \caption{\small Generalization over block numbers in simulation. A successful trajectory is one which all goals are filled at the end.}
    \label{fig:full_comparison_blockstacking}
\end{figure}

\subsection{Comparisons on K-Block stacking}

We compare our trained GNN policy (\textbf{IL-GNN}) against a set of RL baselines on blockstacking environments, designed to highlight the generalization abilities of a GNN policy trained with imitation learning (IL) over RL.
All baselines use the same state and action space as our approach.
Given $g(s_t)$ goals filled at step $t$ and the maximum number of goals to be filled $G$, the dense reward function is $ r(s_t, a_t) = [g(s_{t}) - g(s_{t-1})]/G $ such that a successful $T$-length trajectory has undiscounted return $\sum_{1}^T r(s_t) = 1.0$.

\begin{noindlist}

    \item \textbf{RL-MLP}: RL on an MLP policy; since MLPs have fixed input sizes, we retrain the policy for each stack of size 2-9. %
    
    \item \textbf{RL-GNN}: In this baseline, our GNN policy is trained using RL on stacks of size 2 to 9, and its performance is compared to training with imitation learning.  
    
    \item \textbf{RL-GNN-Seq}: We design this baseline using the sequential training curriculum described in \cite{li2020towards}. 
    The curriculum starts by training our GNN policy for $K_\text{base}=2$ blocks and initializes the policy for $K$ blocks with policy trained in the $K-1$ environment, until $K=9$. This highlights the advantage of IL even over tuned RL training approaches.
    \item \textbf{RL-GNN-demo}: Same as \textbf{RL-GNN-Seq}, but inspired by auxiliary losses from previous work such as \cite{peng2018deepmimic,gupta2019relay,rajeswaran2017learning}, we introduce an additional loss term to the policy optimization.
    Given policy $\pi_\theta$,
    expert demonstrations $\mathcal{D}^\text{Expert} = \{ (s_t^\text{E}, a_t^\text{E}) \}$,
    and cross-entropy loss $\mathcal{H}$,
    the original reinforcement learning loss
    is augmented with an auxiliary loss with hyperparameter weight $\lambda_\text{IL}$:
    $ \mathcal{L}_\text{IL} = \mathbb{E}_{s_t^\text{E}, a_t^\text{E} \sim \mathcal{D}^\text{Expert}}\left[\sum_t \mathcal{H}(\pi_\theta(a_t | s^\text{E}_t), a_t^\text{E}) \right] $.

\end{noindlist}

For all RL baselines, we use Proximal Policy Optimization (PPO) \cite{schulman2017proximal, stable-baselines3} as our training method of choice. 
We give a large environment interaction budget to the RL policies: 2000 environment interactions per stack, resulting in 16,000 interactions in total across $K= 2 \dots 9$. In comparison, our approach \textbf{IL-GNN} is trained on only 90 environment interactions from 20 expert trajectories on box packing and unpacking generated by a simple hand-designed expert policy (Figure \ref{fig:intro}a, \ref{fig:intro}f). We randomize expert trajectories to create augmented dataset of 900 training samples. Due to sample efficiency of the approach (operating at hundreds of environment interactions, not millions), all training converges in less than 20 minutes on a single standard commodity GPU.

As can be seen in Figure \ref{fig:full_comparison_blockstacking}, \textbf{RL-MLP} performs the worst ($0.45 \pm 0.01$ on 3-blocks), and both \textbf{RL-GNN} and \textbf{RL-GNN-Seq} perform better ($0.57 \pm 0.03$ and $0.67 \pm 0.03$ on 3-blocks) at smaller problems. Hence, spatial inductive biases of GNNs improve learning on environments with low numbers of blocks. %
However, the performance of all RL baselines gets significantly worse as the number of blocks increases. For $K \geq 6$, the complexity of the task is too high for RL to learn high-performing policies. Even the \textbf{RL-GNN-demo} baseline, which regularizes the policy to be similar to the same expert data, does not do substantially better outside of the training distribution. In comparison, \textbf{IL-GNN} is trained on expert data of $K=3,4$ blocks, but successfully generalizes to the out-of-distribution 9-block environment (0.85 to 1.0, depending on the GNN architecture).
In further tests of generalization \textit{up to 40 blocks}, the same IL-GNN policies averaged 75\% success, while policies in \cite{li2020towards} trained on $K$ blocks only generalize to $K+1$.

To explore the worse performance of the RL methods, we use GNNExplainer to compare the distribution of the most important features between (1) a GNN policy from the RL-GNN-Seq method, which performs relatively poorly, and (2) a GNN policy trained using our method, which achieves near-optimal behavior. We identify the most salient features each policy paid attention to while making 100 decisions during a 5-blockstacking task. The RL policy (which performs worse) utilizes a less diverse set of features, almost exclusively paying attention to the ``z'' height, while the method trained using our method crucially learns to pay attention to the ``unfilled'' feature along with the height.

\subsection{Generalization to diverse goal configurations}

Once the GNN policy has been trained on an expert dataset of packing and unpacking $K=3,4$ blocks, it is tested on new goal configurations to study generalization to unseen tasks. Note that we use the \textit{same} learned GNN policies for all experiments in this section as in the previous section. 

The \textbf{6-pyramid} experiment tests the policy's ability to achieve \textit{different goal configurations} outside of its training distribution.
All GNN architectures achieve near perfect performance at stacking blocks in a pyramid (Table \ref{tab:generalization}), showing that the policies can generalize to new goals.

In \textbf{3-stack 3-block} the policies generalize to multiple stacks of block and goal positions. Sage and Attention policies are able to solve this task, as well as GCN to some extent, but Gated polices suffer (Table \ref{tab:generalization}). We observed that the gated architecture tends to overfit to small datasets, resulting in poor performance generalization (Section \ref{sec:gated_explainer}).%

\begin{table}
\vspace{0.25cm}
\scriptsize
    \centering
    \begin{tabular}{l c c c c} \toprule
          &  6-Pyramid & 3-block 3-stack & Box rearrangement\\
         \midrule
         GCN       & $1.0 \pm 0.000$  & $0.881 \pm 0.030$ &  $0.585 \pm 0.114$ \\
         Sage      & $1.0 \pm 0.000$  & $0.994 \pm 0.002$ &  $0.955 \pm 0.020$ \\
         Attention & $1.0 \pm 0.000$  & $1.0 \pm 0.000$   & $0.760 \pm 0.081$ \\
         Gated     & $0.95 \pm 0.015$ & $0.6 \pm 0.049$   & $0.201 \pm 0.056$ \\
         \bottomrule
    \end{tabular}
    \caption{\small Generalization results of different GNN architectures on \textbf{blockstacking and box rearrangement tasks in simulation}.}%
    \label{tab:generalization}
\end{table}

In the \textbf{Rearrangement} experiment, the initial state consists of two closed boxes. At this point, the policy is unaware of the location or number of blocks or their goals, simulating a partially observable setting all demonstrations were fully observable. Once the corresponding boxes are opened, the blocks and corresponding goals become observable and are added as object and goal nodes to the GNN. In this setting, we add additional goal nodes to the GNN that capture the empty space on the table. The GNN policy learns to move the covers of both boxes out of the way, by placing them on the table. Next, it swaps the blocks to the empty box, and finally closes both boxes. Trained on demonstrations of opening and closing a single box for packing/unpacking, the policy generalizes to a setting where there are multiple boxes, multiple covers, and multiple storage locations on the table. %
Table \ref{tab:generalization} shows the performance of the different GNN architectures at this task. %
Sage and Attention architectures are able to generalize well ($0.955 \pm 0.020$ and $0.760 \pm 0.081$), but Gated and GCN's performance suffers.   

\subsection{Generalization to hardware}

We also validate our approach by training GNN policies in simulation and applying them to hardware (Fig.\ref{fig:intro}(g-j)) using the Polymetis framework\cite{Polymetis2021}.
We directly deploy Attention policies trained in simulation on hardware, without any fine-tuning.
The GNN policy picks an object and goal from noisy hardware data, and a \prim primitive picks the chosen object and places it in the desired goal location. 

Since the block sizes, robot frame, etc. differ between simulation and hardware, we apply an affine coordinate transform to change the hardware coordinates to simulation. For generalization, it is important that the distribution of block and goal positions are similar between train and test, even if their actual locations are different and noisy. 

We execute 20 runs each of 4 block stacking, 2-stack 3-blocks, 6-pyramid and box packing on hardware, and observe that the trained GNN policy is very robust to hardware disturbances, such as perception noise, picking and placing errors.
Through our extensive real-world experiments (400 real-world \prim movements), we can confidently say that GNN policies trained in simulation can robustly solve blockstacking tasks on hardware, without any fine-tuning needed. 
If the low-level policy fails to place a block in the `right' location, and misses the goal, the GNN policy robustly predicts the next action -- either to replace the misplaced block in the right location, or to place the next block in the correct goal.
The most sensitive feature on hardware is detecting if a goal is filled; this feature can be wrongly detected in the presence of perception noise, and can cause errors in GNN predictions. The position estimation noise can be up to $\approx \pm 2 \text{ cm}$ in our experiments, most significant in the direction pointing away from the camera. 
Table \ref{tab:hardware} summarizes the results of our hardware experiments. In the box packing experiment, the goals are occluded by the bucket; the filled goal feature is incorrectly detected and policy always picks the lowest goal for blocks. In 4-block stacking, the highest goal was misclassified as empty at the end of some runs due to inaccurate perception. Despite these errors, the policies successfully finish all tasks a $100\%$ of the time, showing that GNN policies trained in simulation are robust to spatial noise due to their inductive biases.

\begin{table}
\vspace{0.25cm}
\scriptsize
    \centering
    \begin{tabular}{l c c c c} \toprule
          &  4-Blocks & 6-Pyramid & 3-block 2-stack & Box packing\\
         \midrule
         $\%$ Success & $ 1.00$ & $ 1.00 $ & $1.00 $ & $ 1.00 $ \\
         $\%$ Correct & $ 0.91$ & $1.00 $ & $1.00 $ & $ 0.80$ \\
         \bottomrule
    \end{tabular}
    \caption{\small \textbf{Hardware experiments} on blockstacking and box packing. The robot successfully finishes each task a $100\%$ of the time but sometimes picks the wrong goal due to errors in detection.}
    \label{tab:hardware}
    \vspace{-0.3cm}
\end{table}

\subsection{Dishwasher loading experiments}

\label{sec:dishwasher}

\begin{table}
\scriptsize
    \centering
    \begin{tabular}{l | c c c c}
        \toprule
              \makecell{Scenario } &  6 objects & 8 objects & \makecell{10 objects \\ (training)} & 12 objects \\
         \midrule
             Top/bottom & $0.80 \pm 0.00$ & $0.83 \pm 0.02$ & $1.00 \pm 0.00$ & $0.91 \pm 0.03$ \\
             Left/right & $0.70 \pm 0.03$ & $0.76 \pm 0.02$ & $0.78 \pm 0.05$ & $0.79 \pm 0.03$ \\
         \bottomrule
    \end{tabular}
    \caption{\small \textbf{Dishwasher experiments}: we train using 5 demonstrations of the 5-plate, 5-bowl task and test on 2 target configurations: \textbf{(a)} bowls on top, plates on bottom (Fig. \ref{fig:intro} (l)-(m)), \textbf{(b)} all objects on top, with bowls on right \& plates on left (Fig. \ref{fig:intro} (n)-(o)).}
    \label{tab:dishwasher}
\end{table}

Finally, we apply our method to a more complex task: loading a dishwasher with plates and bowls in different configurations.
We build a dishwasher environment in Habitat Sim \cite{szot2021habitat} using the Replica Synthetic - Apartment 0 dataset (a set of 3D models of an apartment, to be publicly released in the future), with two types of objects (bowls and plates) and a dishwasher with two racks (see Fig.\ref{fig:intro} (k)-(o)).
The training data is created using a game-like interface in a point-and-click manner, where desired dishwasher-loading demonstrations can be easily generated by a layperson.

We demonstrate several types of additional complexity with this experiment: (1) multiple object types (bowls/plates), (2) multiple preconditions for feasibility (two trays, loaded only when pulled out and object-specific desired goals), and (3) a variety of desired configurations specified purely from demonstration (i.e. different ways to load a dishwasher). %
So far, we've only predicted actions over a variable number of objects and goals.
Here, we also predict one of six desired pick and place orientations for objects. Additionally, the GNN chooses between \prim and two other actions of opening/closing both trays.
We formalize this by predicting two additional categorical distributions: one over six discrete desired orientations, and another over three possible dishwasher tray actions (toggle top tray, bottom tray, and a no-op action implying no change in the dishwasher configuration). To summarize, we predict four outputs: (1) which block to pick, (2) which goal to place at, (3) a desired discrete orientation, and (4) whether to toggle either of the dishwasher trays. If the GNN chooses to not change the tray state, \prim is executed.

The training procedure is similar to that of the previous experiments, though we are operating in the extremely low-sample regime and only train on 5 expert demonstrations. Results in Table \ref{tab:dishwasher} show that the trained policies robustly generalize to varied object numbers despite the additional complexities, indicating this method can scale to more difficult environments with very few experiments. We show results on two desired goal configurations; specifying these required just 5 new expert demonstrations in the new setting.

\subsection{Explaining the learned GNN policies}
\label{sec:gated_explainer}
\begin{table}
\vspace{0.25cm}
\scriptsize
    \centering
    \begin{tabular}{l | c c c c} \toprule
         Num. traj &  Attention & Gated & GCN & Sage\\
         \midrule
         5 &  $z$, unfilled & $y$, $z$ & $z$, $z$ & $z$, unfilled\\
         15,000 & $z$, unfilled & $y$, $z$ & $z$, unfilled & $z$, unfilled\\
         \bottomrule
    \end{tabular}
    \caption{\small We compare the top two most \textbf{important features} over different numbers of trajectories used for training and network architectures during a \textbf{3-stack 3-block} task.}
    \label{tab:gnn}
\end{table}
Lastly, we further experiment with GNNExplainer from Section \ref{sec:explainer}. We study the important features for a 3-block 3-stack task and explain the poor generalization performance of Gated GNN policies (Table \ref{tab:generalization}). We train GNN policies on two drastically differing dataset sizes: 5 vs. 15,000 expert trajectories.
A comparison of the most salient features by model are listed in Table \ref{tab:gnn}. 
Gated GNN learns spatial relations which rely on Cartesian positions of the blocks, while the other architectures learn to use the more informative ``unfilled'' feature (GCN only on the larger dataset).

This points to overfitting: the spatial rules the Gated architecture learns may work for the single-stack case and similar single-structure goal configurations such as \textbf{6-pyramid}; however, to rely primarily on $y$- and $z$-features proves insufficient when the goal configuration may contain several different stacks with differing $y$ values, as in the \textbf{3-stack 3-block}.

\section{Conclusion and Future Work}

In this work, we present a graphical policy architecture for manipulation tasks that can be learned with expert demonstrations, and is extremely sample-efficient to train. Once the graph neural network policies are trained, they demonstrate zero-shot generalization behavior across unseen and larger problem instances, along with interpretable explanations for policy decisions.
We test four GNN architectures, finding several that are extremely sample-efficient at learning the underlying structure of the task and generalizing to new tasks. 
Such a GNN policy learned in simulation generalizes well to a real Franka robot.

We hope to address model limitations in future work.
Our method requires pose data of the observed objects and possible goal locations; learning this from raw visual input is a clear next step.
Generalizing to continuous action spaces, for example by directly predicting the 6D desired object pose, will also reduce the reliance on predefined goal positions.
Incorporating past state information into the current timestep's decision, for example by using graph recurrent neural networks \cite{hajiramezanali2019variational}, would allow the policy to adapt to more complex forms of partial observability.
We hope this work opens up exciting avenues for combining research on GNNs with TAMP problems.

\section{Acknowledgements}

{\small We thank Sarah Maria Elisabeth Bechtle Franziska Meier and Dhruv Batra for helpful discussions and feedback on the paper.}

\bibliographystyle{IEEEtran}
\bibliography{references.bib}{}

\end{document}